\documentclass[sigconf]{aamas} 

\usepackage{balance} 
\usepackage{multirow}

\setcopyright{ifaamas}
\acmConference[AAMAS '22]{Proc.\@ of the 21st International Conference
on Autonomous Agents and Multiagent Systems (AAMAS 2022)}{May 9--13, 2022}
{Auckland, New Zealand}{P.~Faliszewski, V.~Mascardi, C.~Pelachaud,
M.E.~Taylor (eds.)}
\copyrightyear{2022}
\acmYear{2022}
\acmDOI{}
\acmPrice{}
\acmISBN{}

\acmSubmissionID{364}

\title[AAMAS-2022 Paper]{Integrated Altruistic and Fairness Preference Induces Advanced Mutual Cooperation in Sequential Social Dilemmas}

\author{Yu Wei}
\affiliation{
  \institution{The University of Tokyo}
  \city{Tokyo}
  \country{Japan}}
\email{weiyu@isi.imi.i.u-tokyo.ac.jp}

\author{Yukiko Ogura}
\affiliation{
  \institution{The University of Tokyo}
  \city{Tokyo}
  \country{Japan}}
\email{ogura@isi.imi.i.u-tokyo.ac.jp}

\author{Yoshiyuki Ohmura}
\affiliation{
  \institution{The University of Tokyo}
  \city{Tokyo}
  \country{Japan}}
\email{ohmura@isi.imi.i.u-tokyo.ac.jp}

\author{Ildefons Magrans de Abril}
\affiliation{
  \institution{The University of Tokyo}
  \city{Tokyo}
  \country{Japan}}
\email{ildefons@isi.imi.i.u-tokyo.ac.jp}

\author{Hoshinori Kanazawa}
\affiliation{
  \institution{The University of Tokyo}
  \city{Tokyo}
  \country{Japan}}
\email{kanazawa@isi.imi.i.u-tokyo.ac.jp}

\author{Yasuo Kuniyoshi}
\affiliation{
  \institution{The University of Tokyo}
  \city{Tokyo}
  \country{Japan}}
\email{kuniyosh@isi.imi.i.u-tokyo.ac.jp}

\begin{abstract}
Inducing cooperation among distributed agents is still a difficult problem in the field of multi-agent reinforcement learning (MARL), particularly in social dilemma situations. There, individual interests are misaligned with the common good and individual rationality leads to suboptimal group outcomes.
In contrast, humans are able to achieve cooperation with one another in such situations. A common explanation for such cooperative behavior is that individuals have social preferences.
In order to achieve cooperation in MARL, we design a new utility function integrating altruistic preferences (incentive for other's reward) and fairness preferences (incentive for equality) from social psychology and behavioral economics, namely, Altruistic and Fairness Preference (AFP), a reward-sharing mechanism which converts one's own and other's rewards to incentives for cooperative behavior.
We performed comparative experiments with standard RL and inequity aversion agents in two challenging sequential social dilemma games, and showed that AFP agents successfully achieved mutual cooperation with more collective rewards and higher equity than the baselines. To further understand the progression of AFP during training, we subsequently explore the effects of altruistic preferences and fairness preferences on agents’ behavior. The results suggest that altruistic preferences encourage agents to contribute to the public goods, and fairness preferences induce mutual behavior between agents.

\end{abstract}

\keywords{multi-agent reinforcement learning, social dilemma, cooperation, altruistic preferences, fairness preferences}

\newcommand{\BibTeX}{\rm B\kern-.05em{\sc i\kern-.025em b}\kern-.08em\TeX}

\begin{document}


\pagestyle{fancy}
\fancyhead{}


\maketitle 


\section{Introduction}

Multi-agent systems can be used to address problems in various applications, such as decentralized control, simulations in economics or sociology, and robotics. Multi-agent reinforcement learning (MARL) is a promising approach for addressing complex tasks without the need for preprogrammed agent behaviors.
Recently, the evolution of cooperation among agents has become an important problem in MARL. Due to the difficulty in designing rewards for multi-agent systems, individual rewards are often intertwined with each other and create unexpected social dilemmas at the group level. This occurs because individual actions are taken to obtain higher individual rewards, and this can lead to suboptimal outcomes for the group \cite{perolat2017multi}. Thus, the collective return obtained by a multi-agent system clearly indicates how well agents can coordinate with each other \cite{jaques2019social}. Decentralized Reinforcement learning agents struggle to learn to cooperate and solve social dilemmas effectively \cite{hughes2018inequity}.

One approach for improving the collective performance in social dilemmas is to directly challenge the reward design problem and establish a methodology for designing individual reward functions. This approach is more directly related to the local observations of individual agents. However, reward design is difficult, even in single-agent cases. Shaping original reward functions is also a complex task, and only a small class of shaped reward functions have been proven to preserve optimality with respect to the true objective function \cite{eck2016potential, sunehag2017value}.

Many studies address these problems using centralized training paradigms \cite{lowe2017multi,foerster2018counterfactual,iqbal2019actor,foerster2018learning}. However, centralized approaches in multi-agent systems do not scale easily to large populations of agents, and the existence of a ``giant'' agent that has unrestricted access to all global information causes the system to be fragile and unrealistic \cite{oroojlooyjadid2019review}. 
Naively optimizing the joint return is also ineffective and is subject to the ``lazy agent'' problem \cite{sunehag2018value}.

An alternative solution is to enable reinforcement learning agents to mimic human cooperative decision-making mechanisms that are well-known in social psychology and behavioral economics. 

LIO focuses on the direct reciprocity mechanism, in which the agents learn how to give incentives directly to other agents using a learned incentive function. This mechanism leads to a near-optimal division of labor in social dilemmas \cite{yang2020learning}.
Jaques et al. \cite{jaques2019social} designed a social influence mechanism to reward agents for having a causal influence over the actions of other agents. Social influence rewards lead to enhanced coordination and meaningful communication protocols.
Hughes et al. \cite{hughes2018inequity} attempted to improve cooperation in social dilemmas by generalizing the inequity aversion model \cite{fehr1999theory} as an intrinsic reward that incorporates a penalty term for the allocation of both advantageous and disadvantageous outcomes among the agents.
Peysakhovich and Lerer \cite{peysakhovich2018prosocial} imbued agents with prosociality by modifying the reward with a weighted sum of one's own reward and that of others. They proved that the reshaped rewards improve the probability of convergence to a good equilibrium in the matrix and Markov stag hunt games. 
McKee et al. \cite{mckee2020social} drew on interdependence theory from social psychology and endowed agents with an intrinsic motivation to prefer certain group reward distributions among self and others. They also explored the effects of diversity in the intrinsic motivation on populations of reinforcement learning agents in mixed-motive Markov games.

Similar to the studies mentioned above, our aim is to induce cooperation to solve social dilemmas in MARL by introduce concepts and models from social psychology and behavioral economics.
Different from previous works which focus on only one type of other-regarding preference, we endow agents with \textit{integrated} altruistic and fairness preferences, which means that agents not only receive incentives as the rewards of other agents increase (as in \cite{peysakhovich2018prosocial} and \cite{mckee2020social}), but they also have an intrinsic motivation to prefer reward equity (as in \cite{hughes2018inequity}). Theoretically, these two universal motives interact with each other \cite{schmidt2011fairness,fehr2003nature}. Thus, we hypothesize that integrating these other-regarding preferences can lead to more opportunities for agents to cooperate.

Additionally, in previous studies, learning to concern inequity aversion was performed using a utility function with rewards obtained up to the current time step. Such backward-looking agents used outcomes rather than predictions to update their policies, causing problems in environments with high stochasticity \cite{peysakhovich2017consequentialist}. In contrast, we avoid these problems by forcing agents to look forward; that is, agents use \textit{expected} rewards to inform their policies in pursuit of improved future rewards for other agents and a more equal distribution of future rewards.

To this end, we introduce altruistic and fairness preference (AFP), a novel, generalized, and forward-looking mechanism for reward sharing. 
To analyze our AFP mechanism, we adopted two games from sequential social dilemma (SSD) game environments. Through experiments, we show that agents trained with AFP learn to cooperate more effectively in these environments. We also explore how altruistic preferences and fairness preferences maintain positive group outcomes, demonstrating that altruistic preferences encourage agents to contribute to the public good and that fairness preferences induce mutual behavior between agents. Thus, the combination of altruistic preferences and fairness preferences promotes mutual cooperation, resulting in a high equilibrium performance in public goods dilemmas.



\section{Background}
\label{Section:background}

\subsection{Partially Observable Markov Games}
We consider MARL in an N-player partially observable Markov game $\mathcal{M}$ defined by the tuple $<S, A, r, T, O>$. The basic Markov game starts with an initial state $s_0 \in \mathcal{S}$, which is determined probabilistically using the initial state distribution. The game is endowed with an observation function $\mathcal{O} : \mathcal{S}\rightarrow \mathbb{R}^d $, which receives $s \in \mathcal{S}$ and outputs incomplete observation information $o$. For each time step $t$, each agent $i$ obtains the observation $o_t$ from the environmental state $s_t \in \mathcal{S}$ and determines the action $a_t \in \mathcal{A}$ to be taken based on the observation. As a result, the game transitions to state $s_{t+1}$ in the next time step $t+1$ based on the state transition distribution $\mathcal{T} (s_{t+1} | s_{t}, a_{t}) : \mathcal{S} \times \mathcal{A} \rightarrow \Delta \mathcal{S}$ in the next time step $t+1$. 

\subsection{Multi-agent Reinforcement Learning}
In Markov games, one machine learning method that is used to help agents learn appropriate policies is deep reinforcement  \cite{mnih2013playing}. In state $s_t \in \mathcal{S}$, each agent $i$ determines action $a_t$ based on policy $\pi$. If the policy is a distribution, the probability that action $a_t$ is sampled is denoted by $\pi(a_t|s_t)$. 
When the agent executes the action $a_t$, the environment probabilistically transitions to the state $s_{t+1} in \mathcal{S}$, and the game environment determines the reward $r^{i}_{t+1}$ (also called the "payoff" in game theory) using the reward function $r(s_t, a_t, s_{t+1})$ and gives it to agent $i$. The agent then evaluates its behavior in the resulting reward time series $\{r_t, r_{t+1}, \cdots , r_T\}$ and updates its polices to obtain more rewards. 
In order to evaluate the policies independently of the states, we assume, given $R_t = \sum^{\infty}_t \gamma^t r_t$ and $V^\pi(s_t) = \mathbb{E}^{\pi}[R_t | s_t]$, that there are two policies, labeled $\pi$ and $\pi^{'}$: 
\begin{equation}\forall s \in \mathcal{S}, V^\pi(s) \geq V^{\pi^{'}}(s),\end{equation}
\begin{equation}\exists s\in \mathcal{S}, V^\pi(s) > V^{\pi^{'}}(s).\end{equation}
When the above equations hold, $\pi$ is better than $\pi^{'}$. The goal of reinforcement learning is to eventually learn optimal policies based on the above definition.

MARL is a natural extension of single-agent reinforcement learning in which multiple cooperative, competitive, or mixed-motive agents work together to accomplish system goals. 
We consider an environment in which observations and actions are distributed across $N$ agents and are represented as $N$-dimensional tuples of local observations in the observation set $\mathcal{O}$ and actions in the action set $\mathcal{A}$. Each agent receives rewards from the environment and knows how much the other agents are rewarded.
Achieving cooperation for system goals is often challenging because of the inherent non-stationary learning environment for each independent learner and the credit assignment problem. The non-stationarity problem is created by multiple agents learning simultaneously \cite{laurent2011world}. In the credit assignment problem, it is arduous to assign proper rewards based on the contribution of each agent in a group. In addition, because the environment is only partially observed from the perspective of a single agent, agents may receive confusing reward signals from the unobserved behavior of their teammates. Therefore, standard decentralized MARL is often unsuccessful because of its inability to explain the rewards a given agent receives. Thus, the paradigm of decentralized training poses difficulties for agents in dealing with the tension between individual and collective returns (i.e., social dilemmas) \cite{yang2020learning}. 

\subsection{Sequential Social Dilemmas}

In the N-player Markov game, each agent's reward depends not only on their own actions and the state of the environment, but also on the actions of other agents. One's own actions influence the experience and payoff of others' actions, and vice versa. The way in which the action set $ \{ a^{1}, \cdots, a^{N} \}$ of agents connects to rewards forms the payoff structure of the game.
In standard decentralized MARL, agents update their policy only to gain as many rewards as possible. Such a setting is similar to the assumption of rationality in classic economics and game theory. The assumption of rational behavior implies that people prefer to make decisions and take actions that deliver more rewards to them rather than actions that are neutral or hinder their interests. Thus, in the payoff structure in which the interests of a group conflict with the interests of individuals, agents would be caught in social dilemmas, as in many classical economic or game-theoretical models, especially mixed-motive matrix games \cite{leibo2017multi}. 

A sequential social dilemma is a spatial and temporal extension of matrix social dilemma games, in which individual selfish actions produce immediate benefits, while their impacts on the collective develop over a longer time period \cite{hughes2018inequity}. These dilemmas are divided into two categories: public goods dilemmas and commons dilemmas \cite{kollock1998social}. In public goods dilemmas, an individual is required to pay the personal cost of providing a resource that is shared by the whole group. In these dilemmas, rational agents are always tempted to get a ``free ride'' and enjoy the benefits of the public resource without paying the individual cost. In commons dilemmas, individuals are guided solely by rationality and deplete the resources shared by all. 



\section{Altruistic and Fairness Preferences}
\label{Section:method}

The principal idea behind this study is to combine multiple elements of human sociality (specifically, altruism and fairness) to encourage agents to coordinate with each other in order to overcome social dilemmas. Moreover, from the perspective of applying artificial intelligence to the real world, the development of sociality for agents is important for giving humans a sense of trust.
Based on the findings of behavioral economics and social psychology on how humans cooperate in social dilemmas, we propose a social reinforcement learning mechanism for reward sharing that applies empirical scientific models of altruism and fairness preferences in order to navigate agents to cooperate in social dilemmas.


\subsection{Altruism: Social Value Orientation}
Altruism is defined as the increase in one's utility with another's payoff. In dilemma situations, many people achieve cooperation by taking into account the interests of others. In contrast to the classical assumption of rationality that an individual considers only his or her own interests, altruism weighs self-interest up against the interests of others. In social psychology, this is called social value orientation (SVO) \cite{au2004measurements, bogaert2008social}. SVO is a factor that influences decision-making in mixed-motive games and has been shown to explain human cooperative behavior in both games and the real world.

Here, we introduce the basic SVO model. Let $r_{self}$ be the reward of an agent received from the environment, and $r_{other}$ be the reward of other agents. The majority of SVOs discussed in the literature can be described using the linear utility function, which is expressed as
\begin{equation}
    u(r_{self}, r_{other}) = \alpha r_{self} + \beta r_{other},
\label{fm:basic_ufunction}
\end{equation}
where the weights $\alpha$ and $\beta$ contain information on the social orientations and the controlling agent's level of altruism \cite{schulz1989recoding}. We constrain $\beta$ and $\alpha$ to be $\beta = 1 - \alpha$ and $\alpha \in [0, 1]$, respectively, as in \cite{peysakhovich2018prosocial}.

\subsection{Fairness: Constant Elasticity of Substitution Utility Function}
Fairness is defined as the intrinsic motivation for agents to avoid reward inequity between themselves and others \cite{fehr1999theory}. To represent the egalitarian orientation of reducing reward inequity, we introduce the constant elasticity of substitution (CES) utility function \cite{arrow1961capital}:
\begin{equation}
    u = (\sum^N_{i-1} a_i r_i^{\frac{\sigma - 1}{\sigma}})^{\frac{\sigma}{\sigma - 1}}.
\end{equation}
This function is frequently used as a utility function in social psychology and behavioral economics, and it has been used to build computational models combining altruistic preferences with inequity aversion  \cite{fehr2006economics,cox2007tractable,andreoni2002giving,sandbu2007fairness}. This function was applied to the SVO utility function, resulting in  
\begin{equation}
    u(r_{self}, r_{other}) = (\alpha r_{self}^{\, \rho} + (1 - \alpha) r_{other}^{\, \rho})^{\frac{1}{\rho}},
\label{fm:social_motive}
\end{equation}
where $u$ is the social utility function, $\alpha \in [0, 1]$ is the selfishness coefficient, and $\rho \in (-\infty , 1]$ is the fairness coefficient. Agents with $\alpha = 1$ are completely selfish, and agents  with $\alpha = 0$ are completely selfless. In addition, agents with $\rho = 0$ do not consider the inequity of rewards, and agents with $\rho \rightarrow -\infty$ pursue completely equal rewards.

Given a group of size $N$, the utility function for agent $i$ is 
\begin{equation}
    u^i_t(r^1_t, \cdots , r^N_t) = (\sum_{j=1}^{N}w^{i}_{j} (r^j_t)^\rho)^{\frac{1}{\rho}}.
\label{fm:basic_u}
\end{equation}
Equation~(\ref{fm:basic_u}) is problematic when the rewards are negative. Thus, we modify Equation~(\ref{fm:basic_u}) to calculate the utility from positive and negative rewards separately via
\begin{equation}
    u_t(r^1_t, \cdots , r^N_t) = (\sum_{j=1}^{N}w^{i}_{j} (\max (r^j_t, 0))^\rho)^{\frac{1}{\rho}} - (\sum_{j=1}^{N}w^{i}_{j} (\max (-r^j_t, 0))^\rho)^{\frac{1}{\rho}}.
\label{fm:minus_u}
\end{equation}

\subsection{Forward View: Shaping Return}

The main problem in Equation~(\ref{fm:minus_u}) for Markov games is that different agents may receive rewards in different time steps. To account for this case in Equation~(\ref{fm:minus_u}), we endow agents with forward views, meaning the agents pursue the improvement of future rewards for other agents and a more equal distribution of future rewards. Here, we adopt a simple form of return shaping \cite{asmuth2008potential}, using returns $ R^i_t = \sum^{\infty}_t \gamma^t r^i_t $ rather than rewards $r^i_t$ to compute utilities:
\begin{equation}
    u_t(R^1_t, \cdots, R^N_t) = (\sum_{j=1}^{N}w^{i}_{j} (\max (R^j_t, 0))^\rho)^{\frac{1}{\rho}} - (\sum_{j=1}^{N}w^{i}_{j} (\max (-R^j_t, 0))^\rho)^{\frac{1}{\rho}}.
\label{fm:utility_fixed}
\end{equation}

As the utility function uses returns as inputs, we apply the utilities to reinforcement learning algorithms as reshaped returns (e.g., Section~\ref{subsection:learning_agents}).


\section{Experimental Setup}
\label{Section:experimental_setup}

\subsection{Sequential Social Dilemma Games}
We carried out several experiments with two SSDs, a public goods dilemma game \textit{Cleanup}, and a commons dilemma game \textit{Harvest}. In each game, all episodes lasted for 1000 steps, and the default environment parameters \cite{hughes2018inequity} were used in this paper to perform our experiments. Note that in Section~\ref{experiment:comparative_evaluation:cleanup}, \textit{Cleanup} is more challenging with a setting that 
agents have a 50\% probability of failing to emit cleaning beam.

\begin{figure*}[htpb]
\centering
\includegraphics[width=0.7\linewidth]{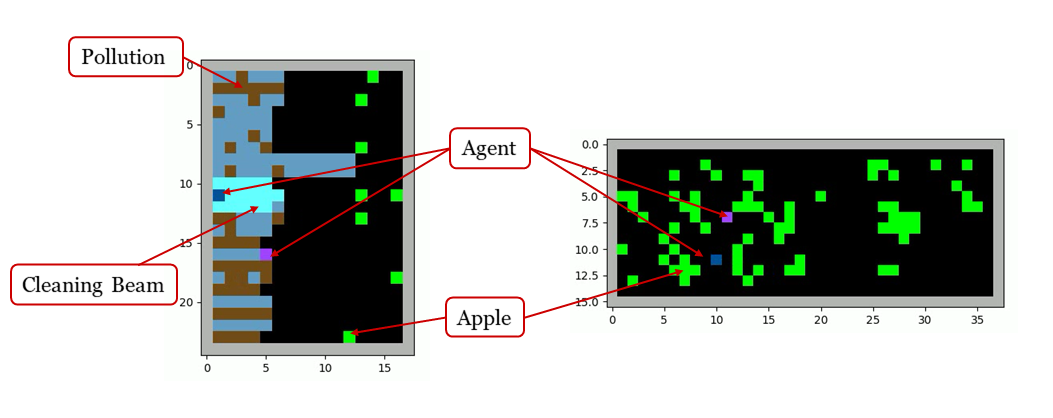}
\caption{Screenshots of gameplay from \textit{Cleanup} and \textit{Harvest}.}
\label{result:screen_shot}
\end{figure*}

\subsubsection{Cleanup}
In \textit{Cleanup} \cite{hughes2018inequity}, players (agents) are rewarded for collecting apples (reward +1) in a 20 x 25 grid world. All players can only observe a 15 x 15 RGB window, which is centered on the agent's position. Apples respawn in an orchard (the black area on the right side of the map) at a rate inversely proportional to the cleanliness of a river (the light blue area on the left side of the map); the cleaner the river, the faster the apples regenerate. The river accumulates pollution over time at a certain rate. When the pollution density exceeds a certain threshold, the response rate of the apples in the orchard drops to zero.
For all agents, the action space consists of "Move", "Clean" pollution, and "Punish" other players (reward -50) at a small cost to the user (reward -1). Thus, players maintain the orchard's regrowth (a public good) with the "Clean'' action and punish free riders with the "Punish" action.

The "Clean" and "Punish" actions act on pollutants and other agents, respectively, by means of a cleaning beam and a penalty beam generated in front of the agents. Because the beams only work within a short distance, a dilemma occurs because in order for apples to spawn, agents must move away from the orchard to clean the river, which confers no reward. A group of agents can sustain apple spawning in the orchard by keeping the pollution density in the river low. However, on shorter time scales, each player would prefer to collect apples in the orchard while the other players clean the river. This creates tension between short-term individual incentives to maximize rewards by staying in the orchard to collect apples and long-term collective incentives to maintain the public good through sustained contributions over time \cite{mckee2020social}.


\subsubsection{Harvest}
In \textit{Harvest} \cite{hughes2018inequity}, the goal is to collect apples (reward +1) in a 24 x 26 grid world. All players can only observe a 15 x 15 RGB window centered on the agent's position. At the beginning of the episode, apples are probabilistically generated on the map. After an apple is collected at a particular point on the map, the apple respawn rate at that point depends on the number of other apples within a radius of 3. If no apple remains in this range, the regrow rate drops to zero. In other words, if agents overharvest apples, apples will be irreversibly depleted. Similar to \textit{Cleanup}, agents in \textit{Harvest} can use the "Punish" action to punish other players (reward -50) at a small cost to the user (reward -1). By using a punishing beam, agents can prevent others from free riding. 'In this game, the dilemma occurs because of the individual temptation to harvest apples quickly, which leads to a lower total yield for the group in the long term \cite{wang2019evolving}.

\subsection{Learning Agents}
\label{subsection:learning_agents}

A distributed advantage actor-critic (A2C) is used as the learning algorithm for our agents to learn policies $\pi$ \cite{mnih2016asynchronous}.
Agents update their policies using the REINFORCE policy-gradient method, employing an advantage function to reduce the variance of the policy gradient. To calculate the advantage function, agents maintain both a value-estimating network as the critic and a policy network as the actor.
In standard A2C, actors calculate their policy gradient for each episode via
\begin{equation}
    \nabla_{\theta} \mathbb{E}_{\tau \sim \pi}[R(\tau)] = \frac{1}{N} \sum_{\tau^{\pi}} \sum_{t=0}^{T-1} \nabla_{\theta} \ln \pi(a_t|s_t) R_t(\tau_t^{\pi}).
\end{equation}
To apply our utility function (Equation~(\ref{fm:minus_u})) in the policy gradient calculation, we substitute the return $R_t(\tau^{\pi})$ with the utility $u_t(\tau^{\pi})$ in the above equation. Therefore, the policy gradient was calculated as 
\begin{equation}
    \nabla_{\theta} \mathbb{E}_{\tau \sim \pi}[u(\tau)] = \frac{1}{N} \sum_{\tau^{\pi}} \sum_{t=0}^{T-1} \nabla_{\theta} \ln \pi(a_t|s_t) u_t(\tau_t^{\pi}).
\end{equation}
Similarly, critics also update their value functions $V(\tau^{\pi})$ with the utility $u_t(\tau^{\pi})$ (instead of the returns) as the update target.
Furthermore, we apply entropy to the loss to encourage agents to explore, as described in \cite{mnih2013playing}.

The neural networks of our agents are all the same; they consist of a convolutional layer, two fully connected layers, a long short-term memory (LSTM) layer, and a linear readout to output the policy $\pi(a_t | s_t)$ and the value function $V(s_t)$. We chose a rectified linear unit (ReLU) as the activation function for all layers. In each time step, agents receive a 15 x 15 egocentric RGB image as an observation. We preprocessed these RGB images into gray-scale images as inputs to the neural networks.

\subsection{Baseline}
The first baseline consists of independent A2C (IAC), which trains agents with standard A2C in the same distributed training framework as our proposed method. We considered two variations in reward signals. In the first variation, labeled IAC-e (egoism), agents receive only their own reward from the environment (i.e., individuals follow the classical assumption of rationality with the goal of maximizing their own interests). In the second variation, labeled IAC-u (utilitarianism), $r^i_t = \frac{1}{n} \sum_i^N r^i_{t, env}$, and agents are aware of the rewards they and others receive and attempt to learn how to maximize the sum total of the rewards for all agents. 

The second baseline consists of inequity aversion (IA) agents \cite{hughes2018inequity}. IA extends the inequity aversion model, a computational model used in behavioral economics as an intrinsic motivation for temporally extended Markov games. Agents have backward views and lose utility when other agents achieve rewards greater or less than one's own at the previous timestep.

All the methods in our experiments used Adam \cite{kingma2014adam} as optimization algorithms and the learning rate $l_r = 1e-4$ after hyperparameter tuning.

\subsection{Social Outcome Metrics}
In single-agent cases, scalar metrics such as the value function or cumulative reward are canonical indicators of the behavioral tendency and performance of agents. In contrast, in multi-agent systems with mixed incentives such as social dilemma games, these scalar metrics can mask many problems of the agent group (e.g., high levels of inequality \cite{mckee2020social}) and can adequately track the state of the system \cite{perolat2017multi}. Thus, we introduce four social outcome metrics to measure system behavior and facilitate its analysis.

The utilitarian metric ($U$) measures the sum of the rewards obtained by all agents in one episode. It is defined as the average sum of individual rewards $R^i$, and it is expressed as
\begin{equation}
U = \dfrac{\sum_{i=1}^N R^i}{T},
\end{equation}
where $T$ is the number of time steps.
The equality metric ($E$) is defined as the inverse of the Gini coefficient:
\begin{equation}
E = 1 - \frac{\sum_{i=1}^N \sum_{j=1}^N \left | R^i - R^j \right |}{2(N-1)\sum_{i=1}^N R^i}.
\end{equation}
This metric measures the inequality in the cumulative rewards of agents in a multi-agent system. The range of the value of $E$ ranges from 0 to 1. The larger the value of the coefficient, the smaller the disparity of the reward in the system. Specifically, $E = 0$ indicates that one agent monopolizes all rewards in the system. Conversely, $E = 1$ means that the rewards obtained by each agent are uniform, and there is no disparity.

The contribution metric ($C$) measures the amount of labor $L$ invested by the entire system to contribute to the public good (i.e., the level of labor of the entire system). The amount of labor $L$ is defined as the average time that the agents take to perform the pre-defined "production activities" $\mathcal{A}_{product}$ in the game environment. For example, in the case of \textit{Cleanup}, $L$ is defined as the ratio of the number of time steps in which the "Clean" action is activated to the total number of time steps in the game. Therefore, $C$ was calculated as the average number of time steps in which labor is performed, expressed as:
\begin{equation}
C = \frac{\sum_{i=1}^N L^i}{T},
\end{equation}
\begin{equation}
L^i = \sum_{t=1}^T I(a_t^i) \ \ \ \ \text{, where }
I(a^i_t) = \left\{\begin{array}{rcl}1 & & \text{if } a^i_t \in \mathcal{A}_{product}\\0 & & \text{otherwise.}\end{array}\right.,
\end{equation}
where $T$ is the number of time steps.

The mutual metric ($M$) measures the disparity in the amount of labor ($L$) invested by agents for the public good:
\begin{equation}
M = 1 - \frac{\sum_{i=1}^N \sum_{j=1}^N \left | L^i - L^j \right |}{2(N-1)\sum_{i=1}^N L^i}.
\end{equation}
It is defined in the same way as the equality metric $E$. In particular, when $M=0$, all labor is provided by a single agent, and the other agents are free riders of the public good produced by this agent. Conversely, when $M=1$, the amount of labor invested by each agent is perfectly uniform.


\section{Results}
\label{Section:result}

\subsection{Comparative Evaluation}

\subsubsection{Cleanup}
\label{experiment:comparative_evaluation:cleanup}
\begin{figure}[htpb]
\centering
\includegraphics[width=1.0\linewidth]{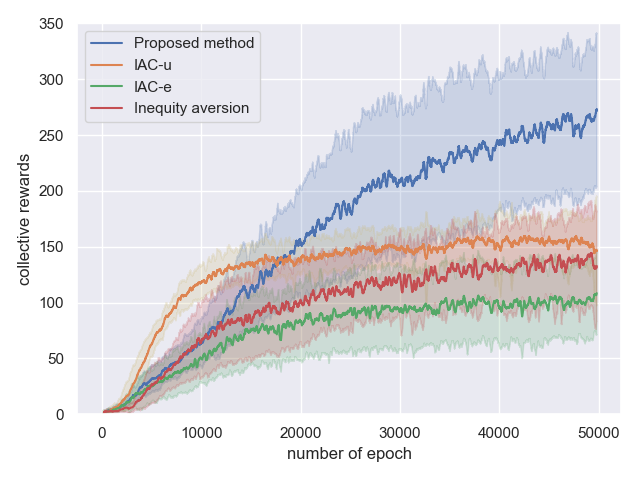}
\caption{Results of learning with limited labor capacity for public goods dilemmas in \textit{Cleanup} with "Hard" setting. The shaded colors represent the standard deviations for the datasets with the corresponding colors.}
\label{result:cleaning_probabilistic_failure}
\end{figure}

We began by training agents to solve \textit{Cleanup} with a group size of $N = 2$.
The selfishness coefficient $\alpha$ and fairness coefficient $\rho$ were selected by conducting an initial hyperparameter sweep. We subsequently chose $alpha = 0.3$ and $\rho = 0.2$ for all AFP agents.
We ran 10 independent trials for all the methods with 10 different random seeds, as well as initializations of the neural networks. Each trial lasted 50000 episodes, and one random seed was used for training. The learning performance of each method was reported as the average return of 10 trials.
Figure~\ref{result:cleaning_probabilistic_failure} shows that the AFP agents collected significantly more rewards than those in the IAC and IA baselines in \textit{Cleanup}. The selfish agents of IAC-e performed the worst. In particular, all the methods generated negative returns at the beginning of training because the policy networks that were initialized with random policies frequently executed the "Punish" action. Subsequently, all the agents quickly updated their policies and learned not to punish one another. However, the baseline agents failed to learn optimal policies to obtain high collective returns. Figure~\ref{result:cleaning_probabilistic_failure} further shows that the learning curve of AFP did not rise at the fastest rate, but it rose until training ended. This suggests that AFP has a relatively slow learning rate, but it results in a better final collective performance than the baselines.

To understand the features of learned policies through all the RL methods we used in the experiments, we investigated the trajectories produced by the policies and learned by the agents and classified the policies at the end of training as a "Cleaner" (labeled as $\pi_C$) or a "Free rider" (labeled as $\pi_F$) based on $N_{clean}$, the average number of times the action "Clean" is performed in one episode. We defined a policy as a "Cleaner" if $N_{clean} \ge 50$ holds or as a "Free rider" if this condition was not satisfied.
For each trial, the joint polices of the two agents formed a policy pair, and theoretically, there were three possible policy pairs in the game: $(\pi_C, \pi_C)$, $(\pi_C, \pi_F)$ or $(\pi_F, \pi_C)$, and $(\pi_F, \pi_F)$. 
Table~\ref{tb:cleanup_joint_policy} shows the converged policy pairs in the 10 trials of all the methods. AFP achieved $(\pi_C, \pi_C)$ in nine out of 10 trials, which is significantly higher than what the IAC and IA baselines achieved. Most of the agents trained by AFP successfully learned to cooperate with each other to clean up waste effectively, resulting in high collective performance. In contrast, most IAC agents discovered a division of labor, whereby one agent cleaned most of the waste in the river while the other agent harvested almost all of the spawned apples. However, the collective reward results for this arrangement indicate that such a joint policy of $(\pi_C, \pi_F)$ is a sub-optimal solution to the game, resulting in a relatively low performance for the baseline methods. 

\begin{table*}[htb]
    \centering
    \caption{Learning results of joint policies in \textit{Cleanup}.}
    \begin{tabular}{c|cc|cc|cc} \hline
    \multirow{2}*{Method} & \multicolumn{2}{|c|}{Number of $(\pi_C, \pi_C)$} & \multicolumn{2}{|c|}{Number of $(\pi_C, \pi_F)$} & \multicolumn{2}{|c|}{Number of $(\pi_F, \pi_F)$} \\
    \cline{2-7}
    ~ & Hard & Normal & Hard & Normal & Hard & Normal \\ \hline
    Proposed method & \textbf{9} & \textbf{6} & 1 & 4 & 0 & 0 \\
    IAC-u & 0 & 0 & 10 & 10 & 0 & 0 \\
    IAC-e & 0 & 0 & 10 & 10 & 0 & 0 \\
    Inequity aversion & 2 & 3 & 7 & 7 & 1 & 0\\\hline
\end{tabular}
\label{tb:cleanup_joint_policy}
\end{table*}




To better capture the complexities of system behavior, we next evaluated the four social outcome metrics for all the methods. Table~\ref{tb:cleanup_probabilistic_som} shows that AFP outperformed IAC and IA in all four metrics. For AFP and IAC, the utilitarian metric ($U$) was stable during different independent trials with different random seeds. In contrast, the performance of the IA agents fluctuated significantly. This demonstrates that the agents of our approach stably achieved cooperation.

\begin{table*}[htb]
    \centering
    \caption{Social outcome metric in \textit{Cleanup}. Standard deviations are displayed in parentheses.}
    \begin{tabular}{c|cc|cc|cc|cc|cc} \hline
    Method & \multicolumn{2}{|c|}{U} & \multicolumn{2}{|c|}{E} & \multicolumn{2}{|c|}{C} & \multicolumn{2}{|c|}{M} & \multicolumn{2}{|c|}{U/C}\\
    \hline
    Proposed method & \textbf{0.31}(0.06) & 0.46(0.04) & \textbf{0.77}(0.11) & \textbf{0.30}(0.03) & \textbf{0.74}(0.26) & \textbf{1.03}\\
    IAC-u & 0.18(0.02) & 0.06(0.10) & 0.28(0.04) & 0.0(0.0) & 0.64\\
    IAC-e & 0.12(0.03) & 0.34(0.05) & 0.18(0.03) & 0.0(0.0) & 0.67\\
    Inequity aversion & 0.14(0.60) & 0.60(0.22) & 0.17(0.06) & 0.22(0.36) & 0.82\\\hline
\end{tabular}
\label{tb:cleanup_probabilistic_som}
\end{table*}

We introduced an additional metric, $U/C$, to measure the efficiency at which the system converted labor into rewards (i.e., the agents' efficiency in using the cleaning beam to keep the pollution level in the river low). Our method achieved not only the highest value of the contribution metric ($C$), but also the highest cleaning efficiency ($U/C$), leading to the best collective performance, as revealed by the utilitarian metric ($U$). 


Moreover, the high value of 0.74 for the mutual metric $M$ of AFP indicates that both agents in the group made adequate contributions to the public good. In contrast, the $M$ for IAC was 0.0, with no fluctuations, demonstrating that in the IAC groups, only one agent, the "cleaner,'' contributed to apple spawning by cleaning the river, and the other agent, the "free rider," specialized in collecting apples and did not clean at all. Therefore, even though the amount of labor ($C$) invested by the IAC-u agents at the group level was almost the same as that of AFP, the cleaning efficiency $U/C$ of IAC-u was approximately 37.86\% lower than that of AFP, resulting in the $U$ of IAC-u being 41.94\% lower than that of AFP. While the $M$ of IA was less than that of AFP but higher than that of IAC, the cleaning efficiency of IA was worse than that of AFP but better than that of IAC.


In terms of the equality of rewards, AFP and IA performed well on the equality metric ($E$), and IAC had relatively low equality. This indicates that AFP and IA guides agents to avoid reward inequalities in the group. AFP tends to reduce reward inequality while encouraging each agent to contribute to the public good. In contrast, IA reduces reward inequality mainly by encouraging the free rider to share a portion of the apples, rather than encouraging more frequent usage of the cleaning beam at the group level.


\subsubsection{Harvest}
Next, we trained agents to solve \textit{Harvest} with a group size of $N = 2$. The selfishness coefficient $\alpha$ and fairness coefficient $\rho$ were selected via an initial hyperparameter search. We subsequentially chose $alpha = 0.3$ and $\rho = 0.6$ for AFP agents and trained all the methods with the same random seed to facilitate the comparison.

\begin{figure}[htbp]
\centering
\includegraphics[width=0.85\linewidth]{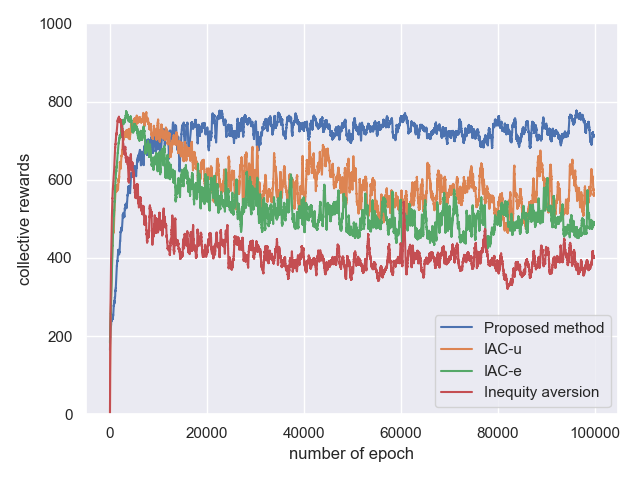}
\caption{Results of learning in commons dilemmas for \textit{Harvest}.}
\label{result:harvest_lr}
\end{figure}

Figure~\ref{result:harvest_lr} shows that AFP agents outperformed those of IAC and IA. The collective reward of our method continuously rose from the start of training to the end, though the learning rate was relatively slow. In contrast, IAC and IA systems moved through two phases. In the first phase, the collective rewards rose dramatically at the beginning and reached a peak. In the second phase, the system performance dropped from the peak and gradually converged to a certain value, which was significantly lower than the performance peak. Additionally, the performance of AFP after its training period had elapsed was similar to the performance peaks of IAC and IA.

As shown in Table~\ref{tb:harvest_som}, we found that all the methods tend to lead to high equality, indicating that equality may be easy to achieve in \textit{Harvest}.

\begin{table}[htb]
    \centering
    \caption{Social outcome metric in \textit{Harvest}.}
    \begin{tabular}{c|c|c} \hline
    Method & U & E \\
    \hline
    Proposed method & \textbf{0.72} & 0.96 \\
    IAC-u & 0.58 & 0.88 \\
    IAC-e & 0.49 & \textbf{0.99} \\
    Inequity aversion & 0.43 & 0.93 \\\hline
\end{tabular}
\label{tb:harvest_som}
\end{table}

\begin{figure}[htbp]
\centering
\includegraphics[width=0.85\linewidth]{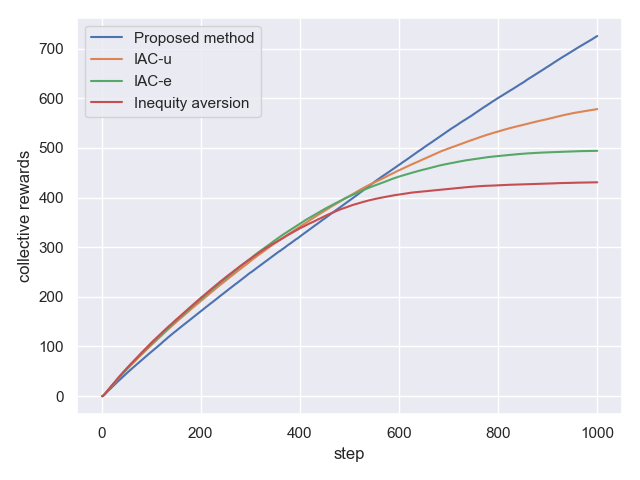}
\caption{Collective rewards as a function of time step for one episode of \textit{Harvest}.}
\label{result:harvest_reward_time}
\end{figure}

We further analyzed the agents' trajectories at the end of training. In Figure~\ref{result:harvest_reward_time}, the reward curve of AFP is almost a straight line, indicating that AFP agents achieved a sustainable growth of collective rewards in a way that did not overexploit the limited public resources. In comparison, the reward curve of IAC and IA agents gradually decreased as the game progressed, indicating that the IAC and IA agents obtained rewards by depleting the apples (i.e., there were hardly any apples left on the map at the end of the game).

\subsection{Social Preferences and Agent Behavior}
How exactly do altruistic preferences (represented by the selfishness coefficient $\alpha$) and fairness preferences (represented by the fairness coefficient $\rho$) affect agents' behavior, especially altruistic behavior? We sought to answer this question by evaluating the effects of altruistic preferences and fairness preferences on agent behavior through the monitoring of social outcome metrics and learning curves.

\subsubsection{Cleanup}
We conducted experiments in \textit{Cleanup} and trained an agent group with a group size of $N=2$. We ran 10 independent trials with different random seeds for each parameter pair of the selfishness and fairness coefficients.

\begin{figure}[htbp]
\centering
\includegraphics[width=1.0\linewidth]{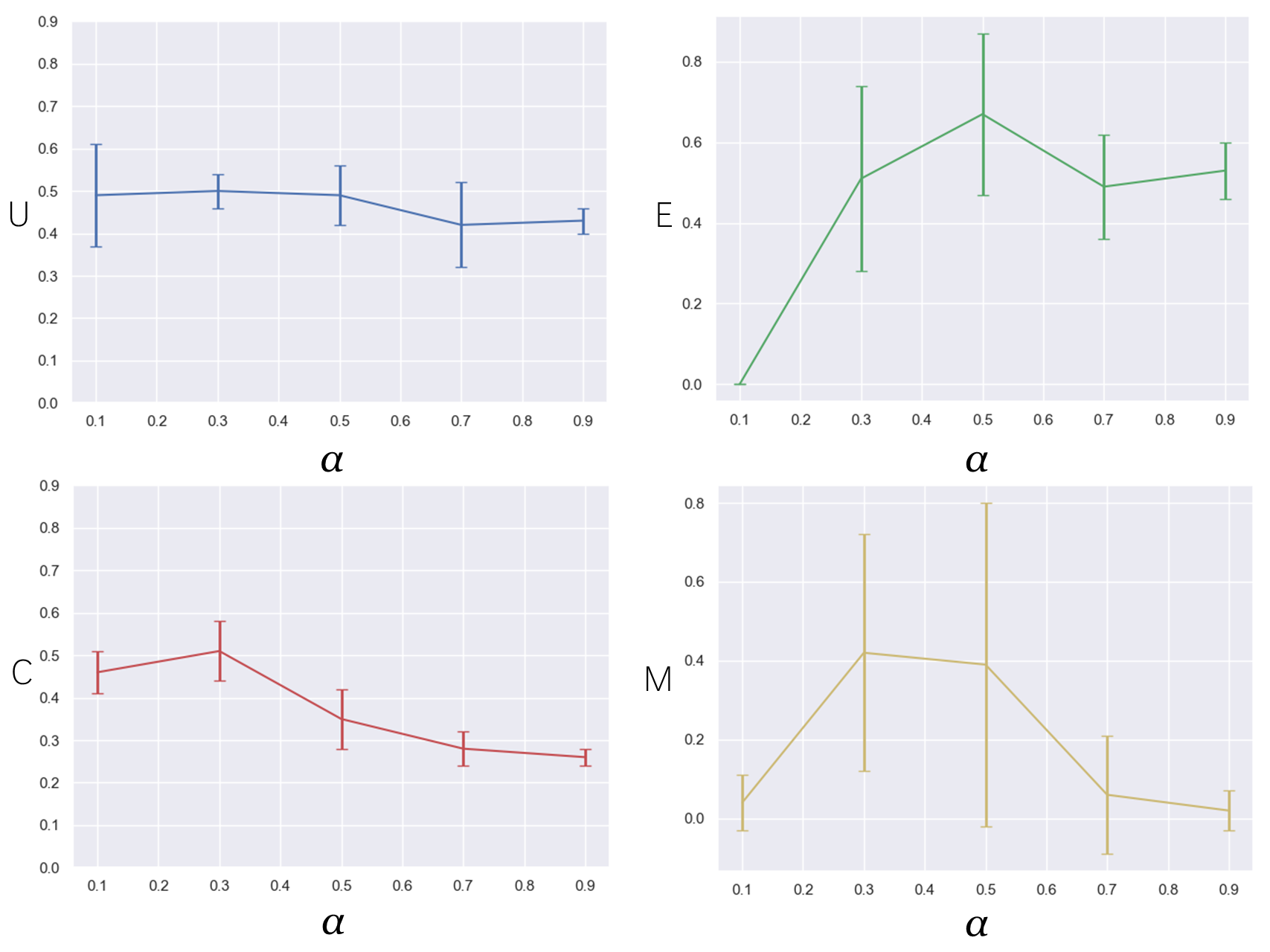}
\caption{Effect of altruistic preferences (selfishness coefficient $\alpha$) on social outcome metrics in \textit{Cleanup}. Standard deviations are displayed as error bars.}
\label{result:cleanup_altruistic_impact}
\end{figure}

In order to evaluate the impact of $\alpha$, $\rho$ was set to 0.6. Figure~\ref{result:cleanup_altruistic_impact} shows the results of the experiment. When $\alpha$ decreased from 0.9 to 0.1, the contribution metric $C$ and the mutual metric $M$ increased, indicating that the level of altruism in the agents' behavior increased when altruistic preferences became stronger. However, when $\alpha$ was extremely small, the equality metric $E$ and the mutual metric $M$ both plummeted.

\begin{figure}[htbp]
\centering
\includegraphics[width=1.0\linewidth]{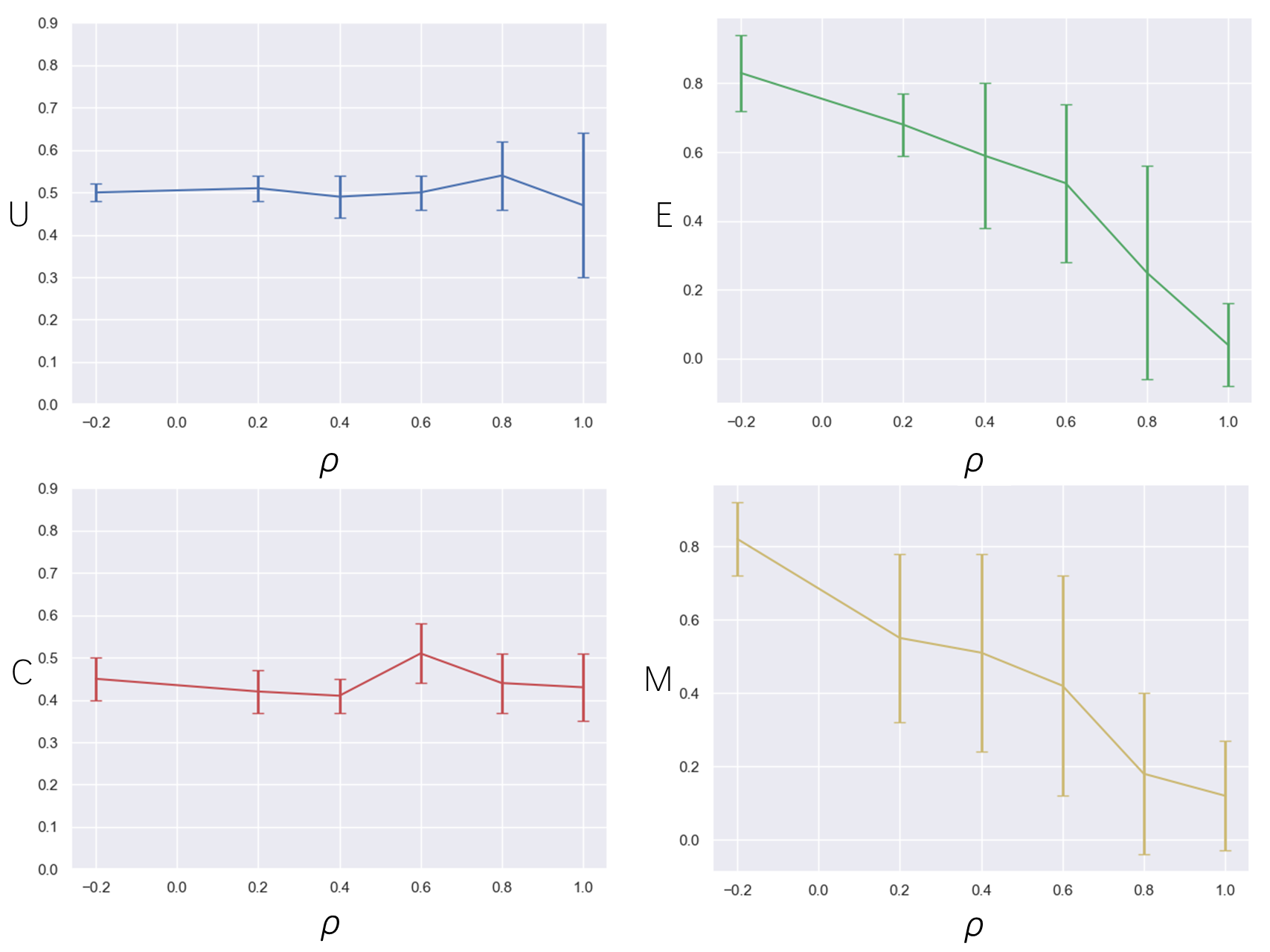}
\caption{Effect of fairness preferences (fairness coefficient $\rho$) on social outcome metrics in \textit{Cleanup}.}
\label{result:cleanup_equity_impact}
\end{figure}

In order to evaluate the impact of $\rho$, $\alpha$ was fixed at 0.3. The experimental results are shown in Figure~\ref{result:cleanup_equity_impact}. When $\rho$ varied from 1.0 to -0.2, there was no significant change in the contribution metric $C$. The smaller $\rho$ was, the larger the equality metric $E$ and the mutual metric $M$ became, indicating that the agents achieved a higher equilibrium performance when the fairness preference for the reward distribution and labor input became stronger.

\subsubsection{Harvest}
In \textit{Harvest}, we studied the impact of the altruistic preferences of trained agents with a group size of $N=2$ and selfishness coefficient $\rho = 0.6$, as well as the impact of the fairness preferences of trained agents with a group size of $N=5$ and selfishness coefficient $\alpha = 0.3$.

\begin{figure}[H]
\centering
\includegraphics[width=0.85\linewidth]{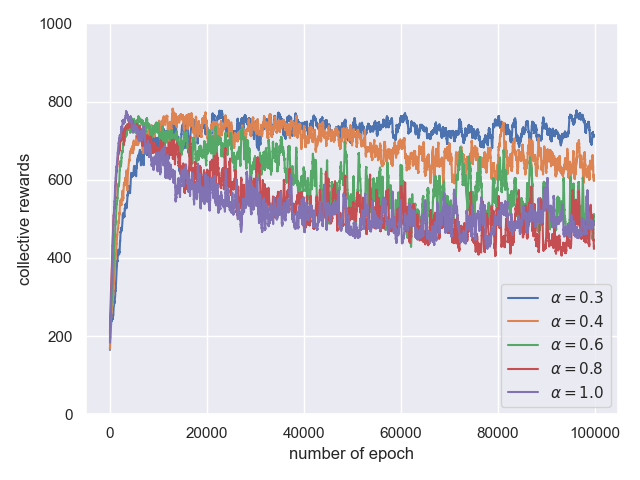}
\caption{Impact of selfishness coefficient on social outcome metric for public goods dilemmas in \textit{Harvest}.}
\label{result:harvest_altruistic_impact}
\end{figure}

\begin{figure}[htbp]
\centering
\includegraphics[width=0.85\linewidth]{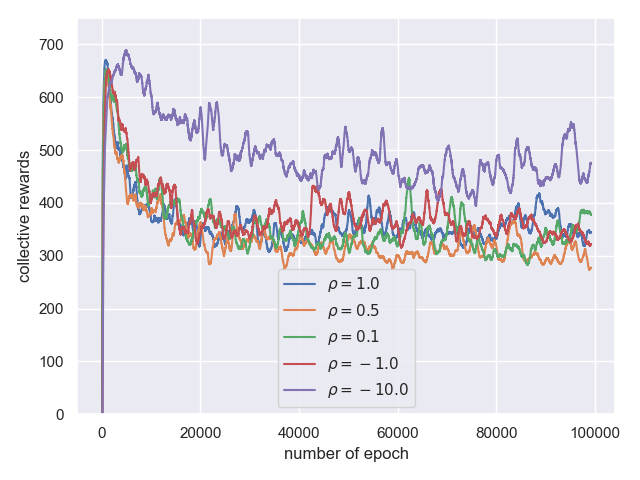}
\caption{Impact of equity coefficient on social outcome metric for public goods dilemmas in \textit{Harvest}.}
\label{result:harvest_equity_impact}
\end{figure}

Figure~\ref{result:harvest_altruistic_impact} shows that performance degradation occurred when $\alpha$ was relatively high. The smaller $\alpha $ was, the smaller the drop in performance. When $\alpha$ decreased to a realistic value, the performance degradation disappeared, indicating that altruistic preferences are effective for resolving commons dilemmas. Figure~\ref{result:harvest_equity_impact} demonstrates that there was no significant difference in performance when $\rho$ varied from 1.0 to -1.0. When $\rho = -10.0$, there was an improvement in performance due to the fairness preferences. However, the performance degradation had not been completely resolved, indicating that fairness preferences are not powerful enough for addressing commons dilemmas in \textit{Harvest}.

\section{Discussion and Conclusion}
\label{Section:discussion}
\balance

Recent studies have explored the possibility of applying knowledge from behavioral economics and social psychology to promote multi-agent cooperation, which is usually difficult for MARL to achieve. We extended the boundary of this exploration by introducing into MARL altruistic and fairness preferences (AFP), a forward-looking reward-sharing mechanism that induces cooperation among decentralized learning agents. In the training process for AFP, agents consider not only their own rewards, but also the rewards of other agents and the equality of rewards among the group. We performed comparative experiments with two baselines (IAC and IA) in two sequential social dilemma games, \textit{Cleanup} and \textit{Harvest}, and showed that AFP agents successfully achieved mutual cooperation with substantial collective rewards and high equity in both environments. In contrast, IAC tended to produce division of labor with hyper-specialized agents, and IA tended to produce inequity-averse agents with low performance.

In \textit{Cleanup}, altruistic preferences encourage systems to invest labor in public goods, but tend to lead to high inequity. On the other hand, fairness preferences induce agents to improve the equality of rewards, and even labor, but it has minimal impact on the amount of labor required at the group level. The combination of altruistic preferences and fairness preferences promotes mutual cooperation in a system, resulting in a high equilibrium performance in such a public goods dilemmas.

In \textit{Harvest}, altruistic preferences encourage agents to passively abstain from overconsuming limited public resources. Sufficient altruistic preferences ensure that agents sustainably exploit public resources. In contrast, fairness preferences are less effective than altruistic preferences because equity can be easily achieved in such an environment.

Our proposed method has several limitations. First, the agent suffers from spurious reward problems because it lacks the ability to explain their rewards. If the rewards of other agents have almost no connection with an agent's own action, then the rewards provide noisy signals for learning and slow down the learning speed. Second, AFP agents tend not to punish free riders in their group, but punishment is a key factor in maintaining cooperation in human society \cite{balliet2011reward}. This greatly reduces the diversity of the agents' policies. Finally, the homogeneity of the agent population was an additional hyperparameter in our settings. Research on reinforcement learning in mixed-motive games has highlighted the fact that policy heterogeneity has a significant impact on agents' behavior. Clearly, this hyperparameter should be properly assigned, especially in games with asymmetric payoff structures.

A natural future direction of our work is to design a mechanism that identifies optimal levels for the altruistic and fairness preferences or that induces agents to explore such hyperparameters automatically. Studying the effects of heterogeneity in the agent population on agents' behavior will also be considered.



\begin{acks}
\end{acks}



\bibliographystyle{ACM-Reference-Format} 
\bibliography{sample}


\end{document}